\documentclass[letterpaper]{article} 
\usepackage{aaai24}  
\usepackage{times}  
\usepackage{helvet}  
\usepackage{courier}  
\usepackage[hyphens]{url}  
\usepackage{graphicx} 
\usepackage{natbib}  
\usepackage{caption} 
\usepackage{algorithm}
\usepackage{algorithmic}

\usepackage{newfloat}
\usepackage{listings}
\DeclareCaptionStyle{ruled}{labelfont=normalfont,labelsep=colon,strut=off} 
\floatstyle{ruled}
\newfloat{listing}{tb}{lst}{}
\floatname{listing}{Listing}
\usepackage{booktabs}

\title{Hypothesis Testing for Class-Conditional Noise\\Using Local Maximum Likelihood}
\author{
    Weisong Yang\textsuperscript{\rm 1}\equalcontrib,
    Rafael Poyiadzi\textsuperscript{\rm 2}\equalcontrib,
    Niall Twomey\textsuperscript{\rm 1},
    Raul Santos Rodriguez\textsuperscript{\rm 1},
}
\affiliations{
    \textsuperscript{\rm 1}University of Bristol,
    \textsuperscript{\rm 2}GSK\\

    \{ws.yang, niall.twomey, enrsr\}@bristol.ac.uk, rafaelpoyiadzi@gmail.com
    
}

\usepackage{bibentry}
\usepackage{amsmath,amssymb,amsfonts}
\usepackage{xfrac}
\usepackage{placeins}

\begin{document}

\maketitle

\begin{abstract}
In supervised learning, automatically assessing the quality of the labels before any learning takes place remains an open research question. In certain particular cases, hypothesis testing procedures have been proposed to assess whether a given instance-label dataset is contaminated with class-conditional label noise, as opposed to uniform label noise. The existing theory builds on the asymptotic properties of the \textit{Maximum Likelihood Estimate} for parametric logistic regression. However, the parametric assumptions on top of which these approaches are constructed are often too strong and unrealistic in practice. To alleviate this problem, in this paper we propose an alternative path by showing how similar procedures can be followed when the underlying model is a product of \textit{Local Maximum Likelihood Estimation} that leads to more flexible nonparametric logistic regression models, which in turn are less susceptible to model misspecification. This different view allows for wider applicability of the tests by offering users access to a richer model class. Similarly to existing works, we assume we have access to anchor points which are provided by the users. We introduce the necessary ingredients for the adaptation of the hypothesis tests to the case of nonparametric logistic regression and empirically compare against the parametric approach presenting both synthetic and real-world case studies and discussing the advantages and limitations of the proposed approach.

\end{abstract}

\section{Introduction}
Data quality checks are an essential part of the machine learning process, as a dataset of poor quality would lead to less trustworthy inferences. There have been plenty of studies targeting different aspects of data quality, e.g. missing value imputation \cite{jarrett2022hyperimpute,lall2022midas}, anomaly detection \cite{xia2022gan,hilal2022financial,schmidl2022anomaly}, or class imbalance \cite{sauber2022use}, etc.

In this work, we focus on assessing the annotation quality of the dataset. In many real-world applications, data annotation is inherently suboptimal, especially when data is harvested from the web \cite{li2017webvision,gong2014multi}, or when the annotation is not carried out by domain experts, such as in the case of crowdsourcing.

Here, our work aims to provide machine learning practitioners with tools for assessing the quality of the labels of a binary supervised dataset of the form: $(\boldsymbol{X}, \boldsymbol{y}) = \{(\boldsymbol{x}_i, y_i)\}_{i=1}^N \in (\mathbb{R}^d\times \{-1, 1\})$. These tools are of the form of hypothesis tests and are meant to take place after the data collection and annotation steps are done. The tests provide information regarding the presence of \textit{class-conditional label noise}, as opposed to \textit{uniform label noise}.
Class-conditional label noise (CCN) refers to the setting where the per-class noise rates are different, i.e. $\mathbb{P}(\tilde{Y} = -1~|~Y=1) = \alpha$ and $\mathbb{P}(\tilde{Y} = 1~|~Y=-1) = \beta$, with $\alpha \neq \beta$. Uniform label noise (UN) refers to the setting where per-class noise rates are identical, i.e. $\mathbb{P}(\tilde{Y} = 1~|~Y=-1) = \mathbb{P}(\tilde{Y} = -1~|~Y=1) = \tau$. 

As other efforts in this area, including those that try to estimate the noise rates directly, a necessary ingredient for our tests is \textit{anchor points}, i.e. instances ($\boldsymbol{x}$), with a known (or, approximately known) \textit{posterior} ($\eta(\boldsymbol{x}) = \mathbb{P}(Y=1~|~X=x)$). 
These need to be provided by the user, probably sourced by a domain expert.


In \cite{poyiadzi2022statistical} the authors introduced a framework that follows from the asymptotic properties of Maximum Likelihood Estimation (MLE) of the logistic regression model and thus its correctness relies on whether the assumptions of the model are met. In practice, the linear assumptions behind the method mean that, unless there is an element of control over the data generating process, the blind application of the test can lead to suboptimal outcomes. In this paper, we take an alternative view by considering more flexible nonparametric models. Instead of relying on the asymptotics of MLE for a parametric model, we build our approach on top of a nonparametric estimation of the underlying regression function based on local likelihood models, such as local polynomial regression, which importantly is more flexible and robust against model misspecification. 

In the empirical analysis, we perform both experiments on synthetic data and real-world data, focusing on understanding the benefits and limitations of the approach as compared to parametric baselines. We show that only if the practitioner knows the data generation process beforehand and can ensure that all the assumptions of the parametric test hold, the parametric test can be used. If any of the assumptions of the parametric test does not hold or if they are unknown at all, we should rely on the nonparametric test, which is more robust against model misspecification. 
We then present a first real-world case study on a smart home dataset. Smart home data can be a typical example that involves dealing with noisy labels, since smart home systems are expected to be scalable and affordable, and most likely non-specialists will be responsible for the data annotation. Consequently, assessing the annotation quality of new datasets will become necessary and important, as it might affect subsequent analyses and conclusions.

The contributions of this paper are summarized as follows:
\begin{itemize}
  \item We extend the parametric form of the hypothesis test for class-conditional label noise proposed in \cite{poyiadzi2022statistical}, and consider a nonparametric estimation of the underlying regression function based on local likelihood models.
  \item We thoroughly compare the strengths and weaknesses of these two methods respectively, and provide guidelines for machine learning practitioners to know which one to use given a dataset.
  \item The effectiveness of our hypothesis test is further illustrated by performing the test on a real-world dataset. We additionally discuss some practical considerations when designing such a test in real life. 
\end{itemize}

\section{Background}

We present here a summary of the main results of local likelihood models that will be used to derive the tests. We first outline some classical results, before introducing kernel-based extensions. Later, we highlight the bias injected by these methods, as well as methods to mitigate these. 

We refer the reader to \citet[Chapter 5]{wasserman2006all} and \citet[Chapter 4]{loader2006local} for a complete presentation. We consider a nonparametric version of logistic regression, with data $(\boldsymbol{X}, \boldsymbol{y}) = \{(\boldsymbol{x}_i, y_i)\}_{i=1}^N \in (\mathbb{R}^d\times \{-1, 1\})$. We want to assess whether the labels have been corrupted with class-conditional noise, as opposed to uniform noise. We assume:
\begin{equation*}
    Y_i \sim Bernoulli\left(r\left(x_i\right)\right)
\end{equation*}
\begin{equation*}
    \hspace{4pt}\textrm{with}\hspace{4pt}r(x_i)= \mathbb{P}(Y_i = 1~|~X_i = x_i) = p_i 
\end{equation*}
for a smooth function $r(x)$ with $0\leq r(x)\leq 1$. Our likelihood function then is:
\begin{equation*}
    \mathcal{L}(r)=\prod_{i=1}^{n} r(x_i)^{Y_i}(1-r(x_i))^{1-Y_i},~~
\end{equation*}
and by using: $\xi(x)=\textrm{logit}(r(x))=\textrm{log}{\frac{r(x)}{1-r(x)}}$, the log-likelihood function:
\begin{equation*}
    \ell(r)=\sum_{i=1}^{n} \ell(Y_i,\xi(x_i))
\end{equation*}
\begin{equation*}
    \hspace{4pt}\textrm{with}\hspace{4pt}\ell(y, \xi) =\log \left[\left(\frac{e^{\xi}}{1+e^{\xi}}\right)^{y}\left(\frac{1}{1+e^{\xi}}\right)^{1-y}\right]
\end{equation*}

To proceed with the estimation of the regression function $r(x)$ near $x$, we approximate the function at $z$ near $x$ with the local logistic function:
\begin{equation*}
    r(z) \approx g\left(\langle \beta, A_p(z-x) \rangle\right),\hspace{6pt}g(v)=\textrm{logit}^{-1}(v)=\frac{\exp(v)}{1+\exp(v)}
\end{equation*}
where $A_p(\cdot)$ is the vector of polynomial basis functions of order $p$. 
For example, $\beta_0 + \beta_1\cdot x$, for $x\in\mathbb{R}^1$. Borrowing the example from \citet[See Eq. 2.9]{loader2006local}:
\begin{equation}
A_2\left(\left[v_0,~v_1\right]\right)~=~[1,~v_0,~v_1,~\frac{1}{2}v_0^2,~v_0v_1,~\frac{1}{2}v_1^2]
\end{equation}

We can now define the local log-likelihood:
\begin{equation}
\ell_{x}(\beta)=\sum_{i=1}^{n} w_{i, h}(x) \ell\left(y_{i},\left\langle \beta, A_p\left(x_{i}-x\right)\right\rangle\right)
\end{equation}
where $w_{i, h}(x)$ is the weight, or the kernel, and is of the form: $w_{i, h}(x) = K\left(\frac{x-x_i}{h}\right)$, where $h$ is called the bandwidth. We use the Gaussian kernel, where: $K(x)=\frac{1}{\sqrt{2 \pi}} e^{-x^{2} / 2}$. See Section 4.2 of \citet{wasserman2006all} for more information on kernels.

Let $\hat{\beta} = \arg\max \ell_{x}(\beta)$. The nonparametric estimate of $r(x)$ is then: $\hat{r}(x) = \frac{\exp{\hat{\beta}_0}}{1+\exp{\hat{\beta}_0}}$.
In the case of a local-likelihood model, we no longer assume a parametric form, as in a (global) likelihood model, but rather fit the polynomial model locally. 

We now discuss common issues and limitations of nonparametric regression models.

\paragraph{The Bias Problem}
\label{sec:bias_problem}
A well-known problem of smoothing methods is a non-vanishing term that introduces bias into the asymptotic normal distribution of the estimand. The implication is that the confidence interval will not be centred around the true function $r(x)$, but rather around: $\bar{r}_n(x) = \mathbb{E}\left[\hat{r}_n(x)\right]$. A few remedies, include: (a) Do nothing, (b) Estimate the bias function: $\bar{r}_n(x) - r(x)$, (c) Undersmooth, and, (d) Bootstrap bias correction. In this work, we follow (a) for illustrative purposes. More information can be found in Section 5.7 of \cite{wasserman2006all}.

\paragraph{Testing for linearity} \cite[Section 5.13]{wasserman2006all}
In this work we introduce an extension of the tests in \cite{poyiadzi2022statistical} that replaces the use of a parametric model, with a nonparametric model. For a practitioner, it would be necessary to know whether a parametric (or linear) fit is suitable for their needs. One option is an F-test \cite{loader_2012} with the null: $\mathcal{H}_0: r(x) = \beta_0 + \beta_1 x$, for $\beta_0, \beta_1$, against the alternative that $\mathcal{H}_0$ is false \cite{wasserman2006all}.

\paragraph{Bandwidth selection (model selection)}
This can be chosen by leave-one-out cross-validation (LOO-CV) as follows,
\begin{equation}
\textrm{LOO-CV}=\sum_{i=1}^{n} \ell\left(Y_{i}, \widehat{\xi}_{(-i)}\left(x_{i}\right)\right)
\end{equation}
In our experiments, we use sub-sampled LOO-CV\footnote{There are also approximations of it as presented in \citet[Chapter 5]{wasserman2006all} and \citet[Definition 4.4]{loader2006local} with describes the Akaike Information Criterion (AIC) adjusted for local likelihood.}. 


\section{Hypothesis Tests based on Local Maximum Likelihood Estimation}
In this section, we introduce the necessary ingredients for the proposed hypothesis tests. These hypothesis tests can be used to provide evidence against the \textit{null hypothesis}: uniform label noise, and for the \textit{alternative hypothesis}: class-conditional label noise. These tests require anchor points to be provided by the user. As we will now discuss, anchor-points are instances for which the true posterior is known to either be strictly, or approximately, $\sfrac{1}{2}$.

We first introduce anchor points and useful identities. We then present properties on the asymptotic distribution of solutions of local maximum likelihood models discussed earlier, and then show how these can be used to construct tests in the cases of having: (1) a single strict anchor-point, (2) multiple strict anchors ($\eta(x_i) = \sfrac{1}{2},~\forall i \in [k],~k>1$, where $k$ is the number of anchor points), (3) multiple relaxed anchors ($\eta(x_i) \approx \sfrac{1}{2},~\forall i \in [k],~k>1$),

\subsection{Anchor points}


In binary classification, we are interested in the posterior predictive distribution: $\eta(\boldsymbol{x}) = \mathbb{P}(Y = 1~|~X=\mathbf{x})$. We will denote the posterior under label noise with: $\tilde{\eta}(\mathbf{x}) = \mathbb{P}(\tilde{Y} = 1~|~X=\mathbf{x})$. Under the two settings: uniform noise $UN(\tau)$ and class-conditional noise $CCN(\alpha, \beta)$, we have\footnote{See of \citet[Section 7.1]{poyiadzi2022statistical} for full derivation.}:
\begin{equation}
    \label{eq:noisy_posterior}
    \tilde{\eta}(\boldsymbol{x})~=\left\{\begin{array}{ll}
(1-\alpha-\beta)\cdot\eta(\boldsymbol{x}) + \beta & \text { if } \textrm{(CCN)} \\
(1-2\tau)\cdot\eta(\boldsymbol{x}) + \tau & \text { if } \textrm{(UN)} \\
\end{array}\right.
\end{equation}

Anchor points are instances for which we are provided with their true posterior. In this work, we are interested in anchor points for which the true posterior is $\sfrac{1}{2}$. Under this setting, we have: 
\begin{equation}
    \eta(\boldsymbol{x})=\sfrac{1}{2}~~~\to~~~\tilde{\eta}(\boldsymbol{x})=\frac{1-\alpha+\beta}{2}
\end{equation}

We will also develop theory around \textit{relaxed} anchor-points (as opposed to \textit{strict} anchor-points discussed above), for which $\eta(\boldsymbol{x}) \approx \sfrac{1}{2}$. More specifically, $\eta(\boldsymbol{x}_i) = \sfrac{1}{2} + \epsilon_i$, for $\epsilon_i \sim \mathbb{U}([-\delta,~\delta])$, with $0 \leq \delta \leq 0.50$, and ideally $0 \leq \delta \ll 1$ (respecting $0 \leq \eta(\boldsymbol{x}) \leq 1$).

\subsection{Asymptotics}
A consistent estimator of the variance of the estimator for a local-likelihood model has the form \cite{frolich2006non}:
\begin{equation*}
    B = \sum_{i=1}^n w_i p_{i} (1-p_{i}) x_{i} x_{i}^{\top}~~~~\&~~~~C\footnotemark = \sum_{i=1}^n\left(y_{i}-p_{i}\right)^{2} x_{i} x_{i}^{\top} w_{i}^{2}
\end{equation*}
\footnotetext{See \citet[Section 2.3]{frolich2006non} for a discussion on similar estimators found in the literature. As noted in said paper, all should result in similar approximations under a small bandwidth.}

\begin{equation}
\hat{\mathbb{V}}[\hat{\beta}_x] = B^{-1}CB^{-1}
\end{equation}
and then with the application of the Delta Method (See for example \citet[Chapter 3]{van2000asymptotic}) we have the variance for the predictions:
\begin{equation}
    \mathbb{V}[\hat{\eta}(x)] \approx \eta(x)^2(1-\eta(x))^2\cdot \hat{\mathbb{V}}[\hat{\beta}_x]_{0, 0}=\frac{1}{16}v(x)
    \label{eq:variance}
\end{equation}

For the covariance we have:
\begin{equation*}
    B_{\boldsymbol{x}_j} = \sum_{i=1}^n w_{i, \boldsymbol{x}_j} p_{i, \boldsymbol{x}_j} (1-p_{i, \boldsymbol{x}_j}) x_{i} x_{i}^{\top}
\end{equation*}
\begin{equation*}
    C_{\boldsymbol{x}_j} = \sum_{i=1}^n\left(y_{i}-p_{i, \boldsymbol{x}_j}\right) x_{i} w_{i, \boldsymbol{x}_j}
\end{equation*}
Note that we now make the independence on $\boldsymbol{x}$ explicit. 
\begin{equation}
    Cov[\hat{\beta}_{x_k}, \hat{\beta}_{x_j}] = B_{\boldsymbol{x}_k}^{-1} C_{\boldsymbol{x}_k}C_{\boldsymbol{x}_j}^{\top}B_{\boldsymbol{x}_j}^{-1}
    \label{eq:covariance}
\end{equation}

\begin{equation}
    cov(\hat{\eta}(x_j), \hat{\eta}(x_k)) = \frac{1}{16}Cov[\hat{\beta}_{x_k}, \hat{\beta}_{x_j}]_{0,0} = \frac{1}{16}cov(x_k, x_j)
\end{equation}

Equations \ref{eq:variance} and \ref{eq:covariance} can be obtained by considering expectations\footnotemark of \citet[Eq. 4.18]{loader2006local}.

\footnotetext{In the form of $\mathbb{E}\left[(X-\mathbb{E}X)^2\right]$ for Eq.\ref{eq:variance} and in the form of $\mathbb{E}\left[(X-\mathbb{E}X)(Y-\mathbb{E}Y)\right]$ for Eq.\ref{eq:covariance}.}

\subsection{A Nonparametric Hypothesis Test for Class-Conditional Label Noise}

We now define our null hypothesis ($\mathcal{H}_0$) and (implicit) alternative hypothesis ($\mathcal{H}_1$) as follows:
\begin{equation}
  \label{eq:htest_noise_rates}
  \mathcal{H}_{0}: \alpha = \beta\hspace{10pt}\&\hspace{10pt}\mathcal{H}_{1}: \alpha \neq \beta
\end{equation}
The next sub-sections outline the varied tests under strict, multiple, and relaxed conditions. 

\subsubsection{Single strict anchors}
Under the null hypothesis, we have the following for the estimated posterior of the anchor:
\begin{align}
    \mathcal{H}_{0}: \hat{\eta}(\boldsymbol{x}) &\sim \mathcal{N}\left(\frac{1}{2},~\frac{1}{16}\cdot v(\boldsymbol{x})\right)\label{eq:htest_null}
\end{align}

\subsubsection{Multiple strict anchors}
Let us consider a set $\mathcal{A}_{\sfrac{1}{2}}^k = \{x~|~\eta(x) = \sfrac{1}{2}\}$, with $|\mathcal{A}_{\sfrac{1}{2}}^k| = k$.
Let $\hat{\eta}_i$ correspond to the $i$th instance in $\mathcal{A}_{\sfrac{1}{2}}^k$. Then for $\bar{\eta} = \frac{1}{k}\sum_{i=1}^k\hat{\eta}_i$ we have:
\begin{equation*}
    \mathbb{E}\bar{\eta} = \mathbb{E}[\frac{1}{k}\sum_{i=1}^k\hat{\eta}_i] = \frac{1}{k}\sum_{i=1}^k\mathbb{E}\hat{\eta}_i = \frac{1}{2}
\end{equation*}

\begin{align*}
    \mathbb{V}\bar{\eta} = \mathbb{V}[\frac{1}{k}\sum_{i=1}^k\hat{\eta}_i] &= \frac{1}{k^2}\left(\sum_{i=1}^k\mathbb{V}\hat{\eta}_i + 2\cdot\sum_{i=1,j>i}^k Cov(\hat{\eta}_i, \hat{\eta}_j) \right)\\
    &= \frac{1}{16k^2}\left(\sum_{i=1}^kv(x_i) + 2\cdot\sum_{i=1,j>i}^k cov(x_k, x_j) \right)\\
    &= \frac{1}{16k^2} var(x_{[1:k]})
\end{align*}

\begin{figure}
    \centering
    \includegraphics[scale=0.30]{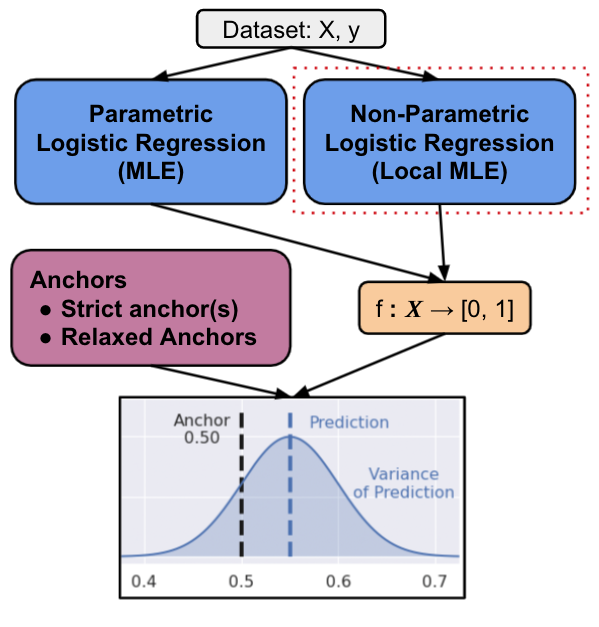}
    \caption{A comparison between the parametric test proposed by \citet{poyiadzi2022statistical} and ours, highlighting that our proposed method offers a richer model class}
    \label{fig:schematic}
\vspace{-8pt}
\end{figure}

\subsubsection{Multiple relaxed anchors}
Providing anchor points with the property that $\eta(x) = \sfrac{1}{2}$ might be challenging, but this strict requirement can be relaxed such that, i.e. $\eta(x) \approx \sfrac{1}{2}$. Following \citet{poyiadzi2022statistical}, we model this with: $\eta(x) = \sfrac{1}{2} + \epsilon$, where $\epsilon \sim \mathbb{U}([-\delta,+\delta$]) ($0 \leq \eta(x) \leq 1$), with $0 \leq \delta \leq 0.50$ denoting the degree of relaxation. We now have two sources of stochasticity: from the model (denoted with the subscript of $\beta$), and from the fact that anchors are not perfect (denoted with a subscript of $A$). We then have:

For one (relaxed) anchor point:
\begin{equation*}
    \mathbb{E}_{\beta}\eta = \frac{1}{2} + \epsilon,~~~~\text{and}~~~~\mathbb{E}_{A}\mathbb{E}_{\beta}\eta = \frac{1}{2}
\end{equation*}

and, with the use of the \textit{law of total variance}:
\begin{align*}
    \mathbb{V}\eta &= \mathbb{E}[\mathbb{V}[\eta|\epsilon]] + \mathbb{V}[\mathbb{E}[\eta|\epsilon]]\\
    &= \mathbb{E}\left[\left(\frac{1}{16} - \frac{\epsilon^2}{2}\right)\cdot v(x)\right] + \mathbb{V}[\frac{1}{2} + \epsilon]\\
    &= \left(\frac{1}{16} - \frac{\delta^2}{6}\right)\cdot v(x) + \frac{\delta^2}{6}
\end{align*}

\noindent and, then for the case of multiple relaxed anchors:
\begin{align}
    \mathbb{V}\bar{\eta} &= \mathbb{E}\left[\left(\frac{1}{16} - \frac{\epsilon^2}{2}\right)\cdot v(x)\right] + \mathbb{V}[\frac{1}{2} + \frac{1}{k}\sum_{i=1}^k\epsilon_i]\nonumber\\
    &= \left(\frac{1}{k^2}\right)\left(\frac{1}{16} - \frac{\delta^2}{6}\right)\cdot var(x_{[1:k]}) + \frac{\delta^2}{6k}\nonumber\\
    &\approx \frac{1}{16k^2} var(x_{[1:k]})\label{eq:var_multiple_relaxed}
\end{align}
where the approximation at the last line follows from the assumption that $\delta$ is small.


\subsection{Parametric vs. Nonparametric Approaches}

As seen by the schematic diagram in Figure \ref{fig:schematic}, in this work we propose a nonparametric alternative to the framework proposed in \cite{poyiadzi2022statistical} under nonparametric logistic regression models, as opposed to only parametric regression models. Therefore, the comparison of the two works can be discussed on the basis of parametric versus nonparametric generalised linear models \cite{wasserman2006all}. In Table~\ref{table:comparison} we summarise the main differences between both approaches.

\begin{table}
    \begin{tabular}{  l  p{5cm}}
        \toprule
        \textbf{Topics}      
        & \textbf{Discussion}\\\midrule
        Flexibility 
        & Nonparametric models are a richer model class than parametric models, and are more robust to model misspecification. This comes at a cost of slower convergence \cite{wasserman2006all}.\\\hline
        Model Fitting       
        & Bandwidth selection be more computationally demanding in the nonparametric case, although approximations exist.\\\hline
        Bias Problem       
        & Smoothing methods are in general biased. See Section Background\ref{sec:bias_problem} for discussion, references, and remedies\\
        \bottomrule
    \end{tabular}
    \caption{Differences between parametric and nonparametric approaches.}
    \label{table:comparison}
\end{table}

\section{Related Work}
Being investigated extensively in the literature, learning with noisy labels \cite{review} has recently had two dominant research directions: (a) direct estimation of noise \cite{cheng2022instance,zhu2022beyond,liu2023identifiability}, and (b) data separation \cite{key2023statistical,karim2022unicon}.  

Direct noise estimation often involves the notion of the noise transition matrix $T(x)$, which reveals the transition probability of clean labels to noisy labels given an instance, i.e. its (i, j) entry $T_{ij}(x)$ is defined as $\mathbb{P}(\tilde{Y} = j~|~Y=i, X=\mathbf{x})$. The transition matrix usually needs to be estimated from the whole noisy dataset in a supervised way and typically requires a large number of training samples, and then it can be used to correct the training loss. However, $T(x)$, in reality, is hard to estimate \cite{cheng2022instance,li2022improving}, especially in scenarios with a high noise rate.

Data separation methods, on the other hand, work on the identification of noisy or clean examples. Typically, this line of research focuses on first filtering out the noisy examples from the clean ones, e.g. using the small-loss trick \cite{yu2019does}. Clean labels are then used to train the model under a supervised setting. In the meantime, noisy labels can be treated as unlabelled data and different semi-supervised learning techniques can be applied \cite{li2020dividemix,chen2023two,han2018co,yu2019does}. Additionally, some data separation techniques are more data-centric as opposed to learning-centric \cite{zhu2022detecting,wei2022self}. A recent study \cite{key2023statistical} proposes a nonparametric hypothesis test to filter out mislabelled instances from correctly labelled ones. The proposed hypothesis test is based on the distance between instances, with the primary assumption being that distances between instances within the same class are stochastically smaller than distances between instances from different classes. To detect mislabelled instances, this procedure is applied to all points in each class, and can be iterated across all classes. One possible limitation of this approach may be the case when there is a large number of mislabelled instances, which could bias the estimates. In addition, how to draw an accurate line between noisy examples and examples with large losses has always been a challenge \cite{wei2022self,pleiss2020identifying}

Interestingly, by looking at these two lines of research from a causal perspective, the authors in \cite{yao2023better} investigate the influence of the data generation process (X causes Y or Y causes X) on the algorithm design, and introduce an intuitive method for the causal structure discovery.

By contrast, the nonparametric hypothesis tests we propose in this work build upon both categories but fundamentally differ on their purpose, as our tests are designed to be employed before any learning takes place -- they are a tool to assess the validity of the annotations of a given dataset (similarly to \cite{poyiadzi2022statistical}). However, different to existing tests, our tests are nonparametric and are based on the solid asymptotic properties of local maximum likelihood estimation.
\section{Experiments}
The empirical analysis is designed to provide a clear understanding of the advantages and disadvantages of nonparametric vs parametric tests in order to provide practitioners with intuition for their application in practice\footnote{We make our code available on GitHub. \url{https://github.com/weisongyang/htest}}. To achieve this, we first explore controlled synthetic datasets and then move to a real-world scenario.

\subsection{Synthetic datasets}

\subsubsection{Symmetric XOR dataset}
Firstly, we consider a synthetic XOR dataset with Gaussian distributions centred at: $[2, 2], [-2, -2], [-2, 2]$ and $[2, -2]$ with scale $1$ (first two correspond to one class, and the latter two to another). Anchor points can be generated by first considering the Bayes Classifier (since we know the data generating distribution), and then optimising the input (the instance) such that the output is $0.50$. In the case of wanting relaxed anchor points, we do it with rejection sampling. In both cases, anchor points are restricted within $[-4, 4]$ for both features. We vary the following parameters: 
\begin{enumerate}
    \item \label{enum:n} $N \in [200,~500,~1000]$: the training sample size.
    \item $(\alpha,~\beta)$: we let $\alpha = 0.0$, and $\beta = 0.10$. Please see appendix for additional experiments on more combinations of $(\alpha,~\beta)$.\label{enum:ab}
    \item $k \in [1,~2,~4,~8,~16]$: the number of anchor points.\label{enum:k}
    \item $\delta \in [0,~0.05,~0.10,~0.20]$: how relaxed the anchor points are: $\eta(x) \in [0.50-\delta,~0.50+\delta]$.\label{enum:delta}
\end{enumerate}
For each combination of $N$ and $(\alpha,~\beta)$ we perform $100$ runs from $100$ new dataset draws. Then for each model, and for each combination of $k$ and $\delta$ we do $10$ draws.

\begin{figure}
    \centering
    \includegraphics[width=\linewidth]{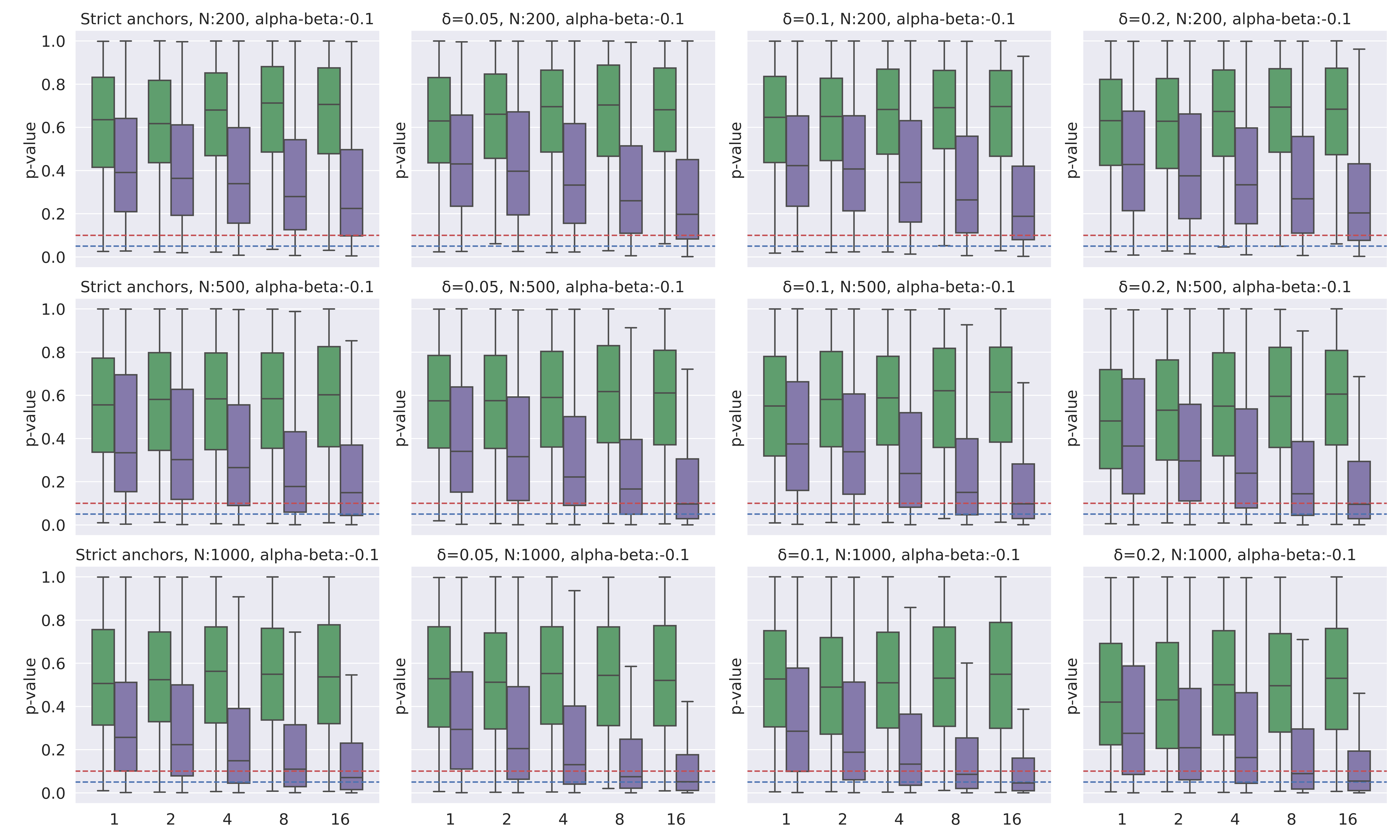}
    \caption{Nonparametric test on symmetric XOR data. Box-plots with $\alpha=0.0$ and $\beta=0.10$ (purple) against box-plots with $\alpha=0.0$ and $\beta=0.0$ (green).}
    \label{fig:nonparametric_alpha0.0_originalxor}
\end{figure}

The sub-figures in Figures \ref{fig:nonparametric_alpha0.0_originalxor} are related as follows: moving from top to bottom, we increase $N$: $200,~500,~1000$, moving left to right we relax the anchors more: Strict, $\delta=0.05, \delta=0.10, \delta=0.20$. Within each sub-figure, we have the p values for different sizes of sets of anchor points: $1,~2,~4,~8,~16$. Then, for every set of anchor points, we have two box-plots, the purple corresponds to the underlying model being trained on noisy data, and the green corresponds to the underlying models being trained on clean data. For example, the sub-figure at the bottom right of figure \ref{fig:nonparametric_alpha0.0_originalxor} refers to training with $N=1000$ data points, anchor points being relaxed with $\delta=0.20$. In all subplots, we indicate 0.10 with a red dashed line and 0.05 with a blue dashed line, which serves as the rejection threshold for the null hypothesis.



From Figure \ref{fig:nonparametric_alpha0.0_originalxor}, we observe that: (1) as the sample size ($N$) increases, (2) as the number of anchor points increases, p values correctly decrease in value implying evidence against the null hypothesis (uniform label noise). We also observe that (per row) moving from right to left, i.e. from the most relaxed to strict anchor points, we again see better performance. In the appendices, we show that as the difference between $\alpha$ and $\beta$ increases, p values correctly decrease as well.

\begin{figure}
    \centering
    \includegraphics[width=\linewidth]{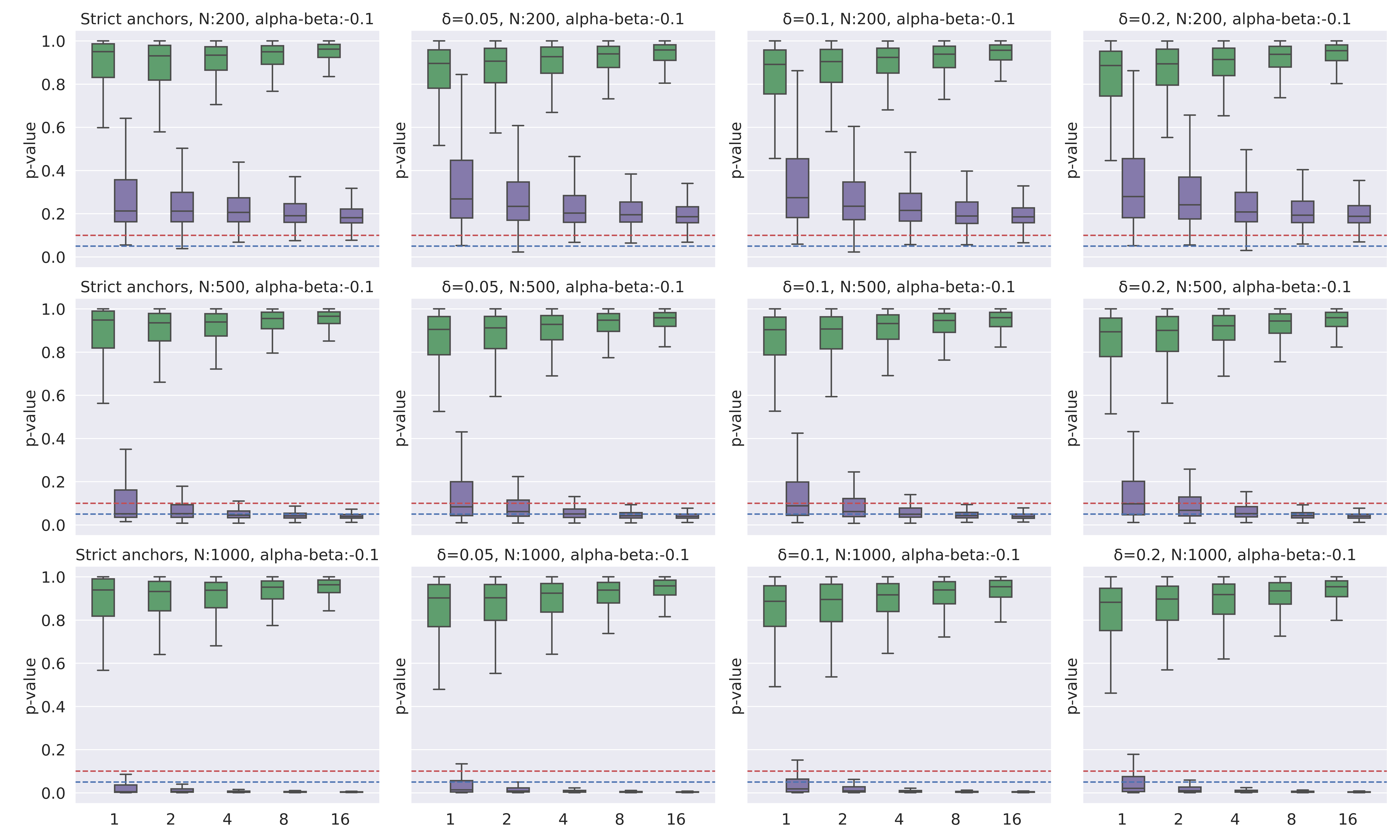}
    \caption{Parametric test on symmetric XOR data. Box-plots with $\alpha=0.0$ and $\beta=0.10$ (purple) against box-plots with $\alpha=0.0$ and $\beta=0.0$ (green).}
    \label{fig:parametric_alpha0.0_originalxor2}
\end{figure}

In addition, we implement the parametric test proposed in \citet{poyiadzi2022statistical} and also run it on the symmetric XOR data. From Figure \ref{fig:parametric_alpha0.0_originalxor2}, we can observe that as the number of training samples and the number of anchor points increase, green boxes go up to 1 and purple boxes decrease to 0, which suggests that the parametric test also works on the symmetric XOR even if its model assumptions are not met. This can be explained by the symmetry and overlap of the symmetric XOR. Thus, we extend the experiment to an asymmetric XOR.

\subsubsection{Asymmetric XOR dataset}
The asymmetric XOR dataset also is consisted of also four Gaussians, which are centred at $[4, 4], [-2, -2], [-1, 1]$ and $[1, -1]$. Apart from this, the rest of the experiment settings are the same as in the symmetric XOR's case.

\begin{figure}
    \centering
    \includegraphics[width=\linewidth]{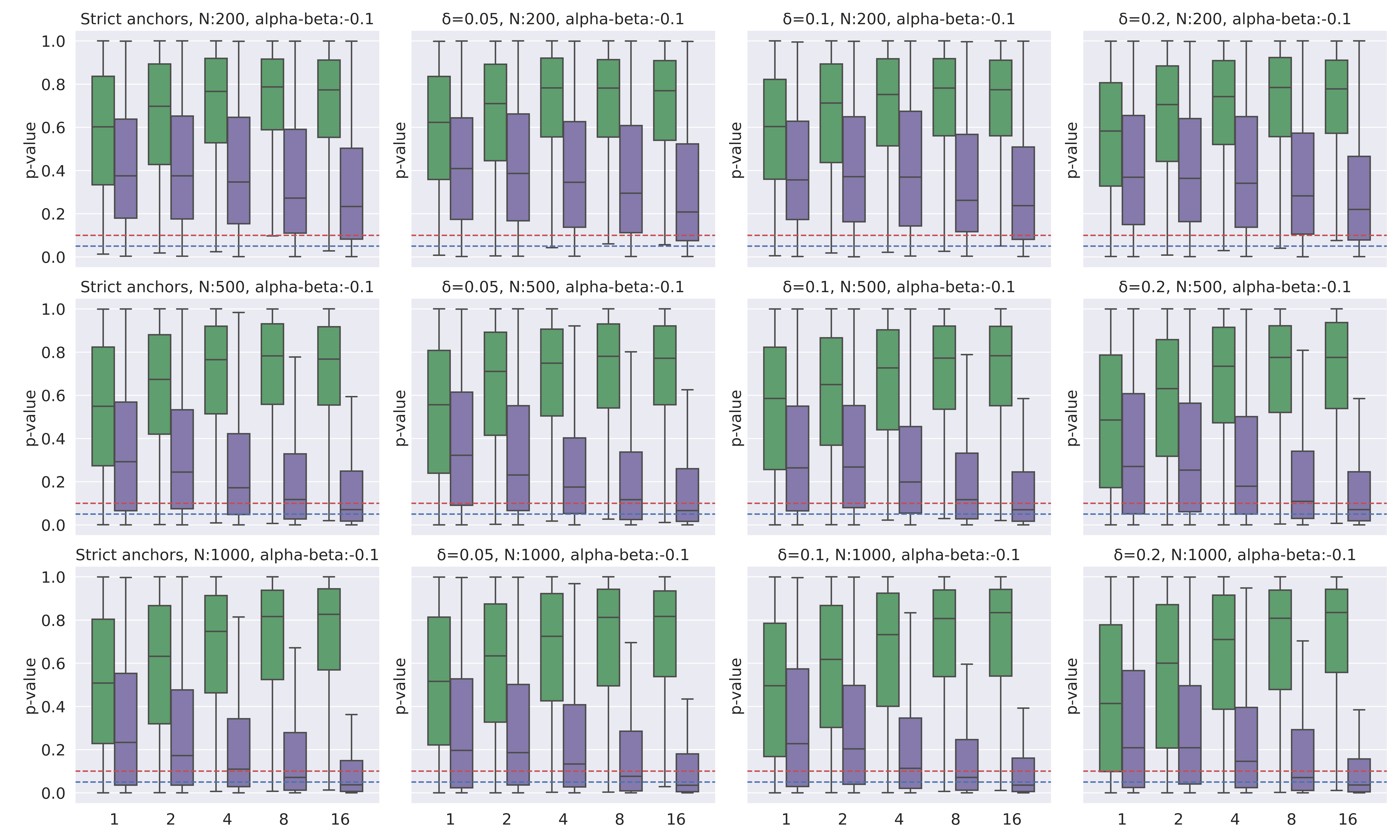}
    \caption{Nonparametric test on asymmetric XOR data. Box-plots with $\alpha=0.0$ and $\beta=0.10$ (purple) against box-plots with $\alpha=0.0$ and $\beta=0.0$ (green).}
    \label{fig:nonparametric_alpha0.0_d4xor2}
\end{figure}

\begin{figure}
    \centering
    \includegraphics[width=\linewidth]{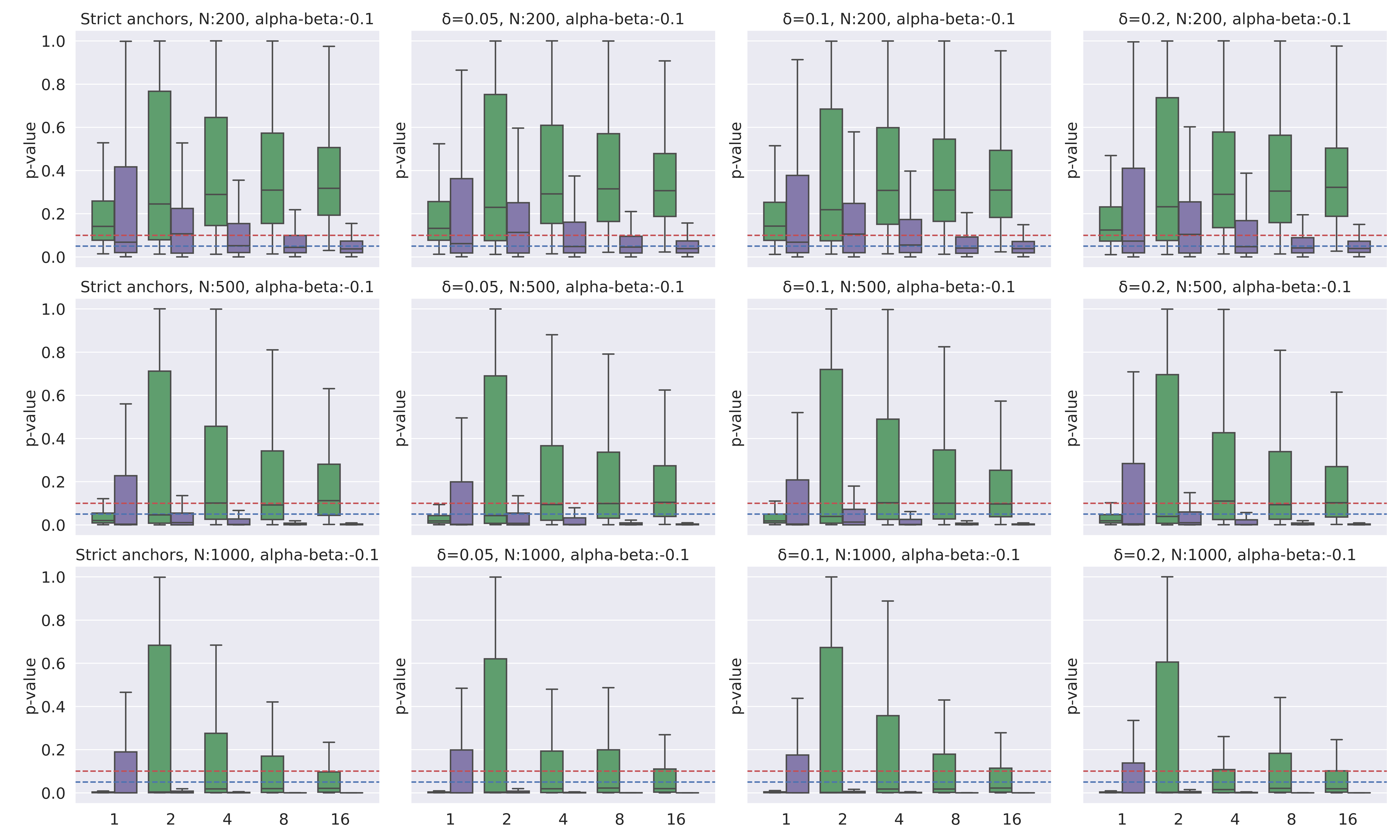}
    \caption{Parametric test on asymmetric XOR data. Box-plots with $\alpha=0.0$ and $\beta=0.10$ (purple) against box-plots with $\alpha=0.0$ and $\beta=0.0$ (green).}
    \label{fig:parametric_alpha0.0_d4xor2}
\end{figure}

For the nonparametric test, from Figure \ref{fig:nonparametric_alpha0.0_d4xor2}, we can make the same observations as in Figure \ref{fig:nonparametric_alpha0.0_originalxor}, i.e. p values will decrease as the training sample size and the number of anchor points increase. Please see the appendices for the results of more combinations of $(\alpha,~\beta)$.

As for the results of the parametric test on the asymmetric XOR, however, we can now notice from Figure \ref{fig:parametric_alpha0.0_d4xor2} that both green and purple boxes will go down as the training sample size and the number of anchor points increase. This verifies our claim that blind application of the parametric test can lead to suboptimal outcomes if any of the model assumptions doesn't hold.

\subsubsection{Limitations} One open question when working with local likelihood problems is the empty-neighbourhood problem, which is more apparent in the low-data setting, and when the instance we are fitting for is rather far from the data. In the future, we would like to better understand remedies for this for our case. Overall the experiments carried out, we have encountered an issue of approximately $1/10000$.



\subsection{Real-world dataset}
\subsubsection{Dataset description}
For the experiment on real-world data, we consider data routinely collected in smart homes. There are numerous initiatives that work on digital health technologies to monitor the health status of people living in smart homes \cite{brumitt2000easyliving,cook2006health,zhu2015bridging,woznowski2017sphere,twomey2016sphere}. With our rapidly growing ageing population and relatively few clinicians, there is a pressing need to automatically understand the state and progression of chronic diseases. The emergence of state-of-the-art sensing platforms offers unprecedented opportunities for indirect and automatic evaluation of disease state through the lens of behavioural monitoring. One of the end goals of the research and development of these smart home systems will be to make them affordable and scalable, so that they could be installed by non-specialists and even general older adults. Naturally, the need to evaluate the quality of data annotation upon installation will emerge. In these settings, a particularly important task is that of localisation which is helpful for quantifying behaviours in the home environment \cite{d2012indoor,otsason2005accurate,klingbeil2008wireless}. Our real-world data is collected from a similar smart home project \cite{weisong2022}. For the localisation task, we have access to room-level annotations. The experiments we carry out in this section serve as a case study to show how hypothesis tests can be used in real life and the practical considerations needed.
\begin{figure*}
    \centering
    \includegraphics[width=0.71\linewidth]{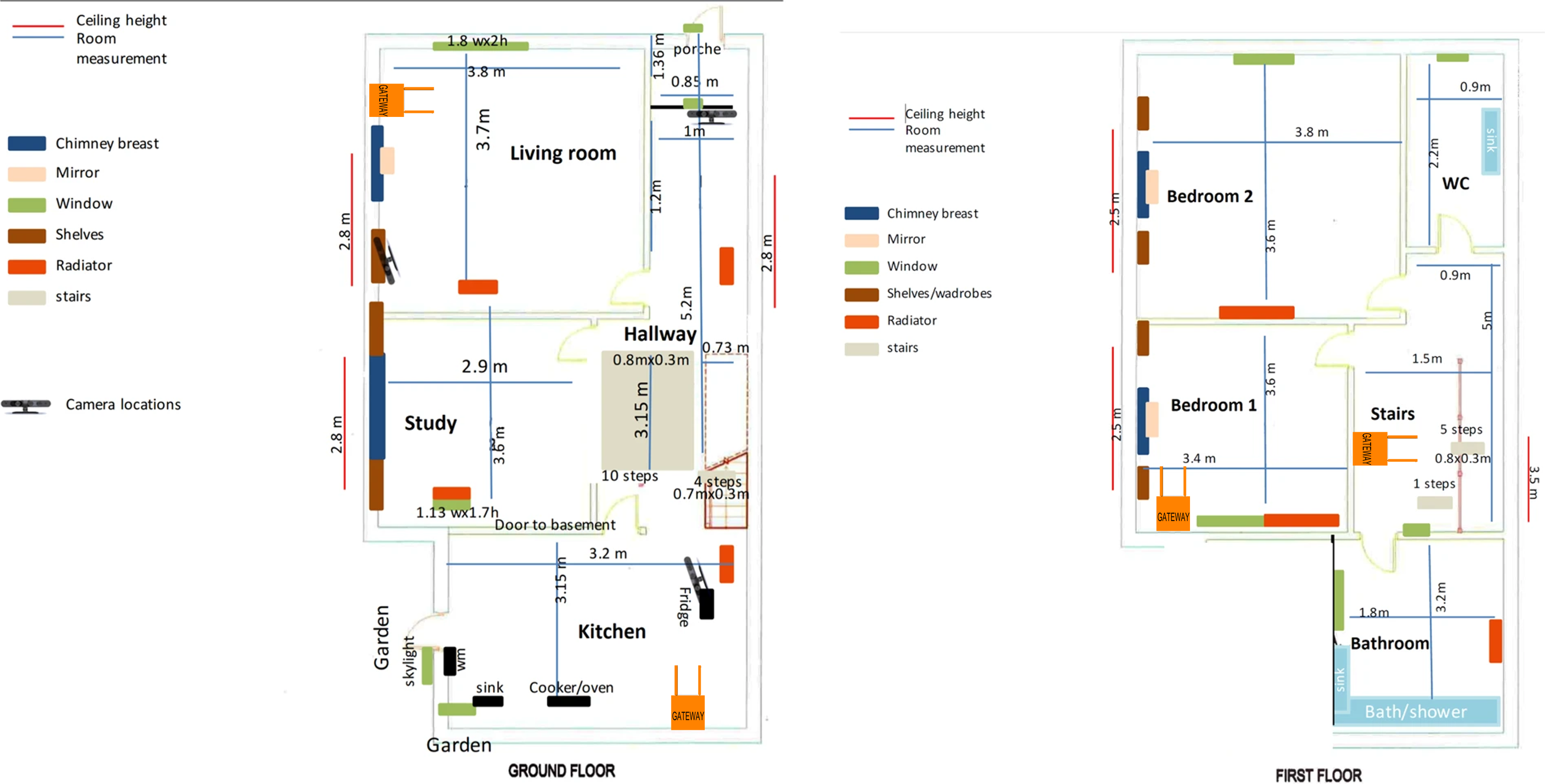}
    \caption{An example house with two floors.}
    \label{fig:floorplan}
\end{figure*}

\subsubsection{Experiment settings}
There are four gateways installed at different locations in our example house, i.e. Living room, Kitchen, Bedroom 1, and Stairs (on the first floor). Figure \ref{fig:floorplan} is its floorplan. These gateways simultaneously record accelerometer data and the Received Signal Strength Intensity (RSSI), which is the basis for our localisation efforts. To be consistent with the binary classification setting in this paper, we group our room-level labels by floor so that we have two classes in this two-floor house. Location annotations were collected by the technician upon installation, who carried all the wearables in a bag and walked through every room. The real-time location of the technician was recorded with an annotation app while the RSSI data is simultaneously recorded. For this example house, the annotation walkaround lasted less than 30 minutes. For input, we extract basic features from windowed RSSI data ($5$ seconds length with an overlap of $2.50$ seconds following \citet{twomey2018comprehensive}). There are two wearables for this house, which gives us roughly 1200 data points. We randomly sample 1000 data points to train our model in each run, and we perform 100 runs in total.

We use RSSI readings received by the gateways when the technician was on the stairs as \textit{anchor points}. The intuition is that the stairs are in the middle of two floors, and RSSI readings on the stairs should be close to the decision boundary of the binary classification task. In other words, the true posteriors of RSSI readings on stairs should be close to 0.5. We vary the number of anchor points $k \in [1,~2,~4,~8]$. Importantly, the notion of stairs also requires extra care when designing the experiment. Depending on the height of the ceiling, the width of the stairs, and on which floor the gateway labelled 'stairs' is installed, the posterior of steps at the bottom of the stairs and at the top of the stairs can be very different, which may introduce bias if we group stairs into one of the classes and include them in the training data. Accordingly, we consider two settings in the end: (a) include RSSI readings on stairs in the training process and use them as anchor points as well, (b) exclude RSSI readings on stairs from the training process, and only use them as anchor points.

\subsubsection{Results}

\begin{figure}
    \centering
    \includegraphics[width=0.9\linewidth]{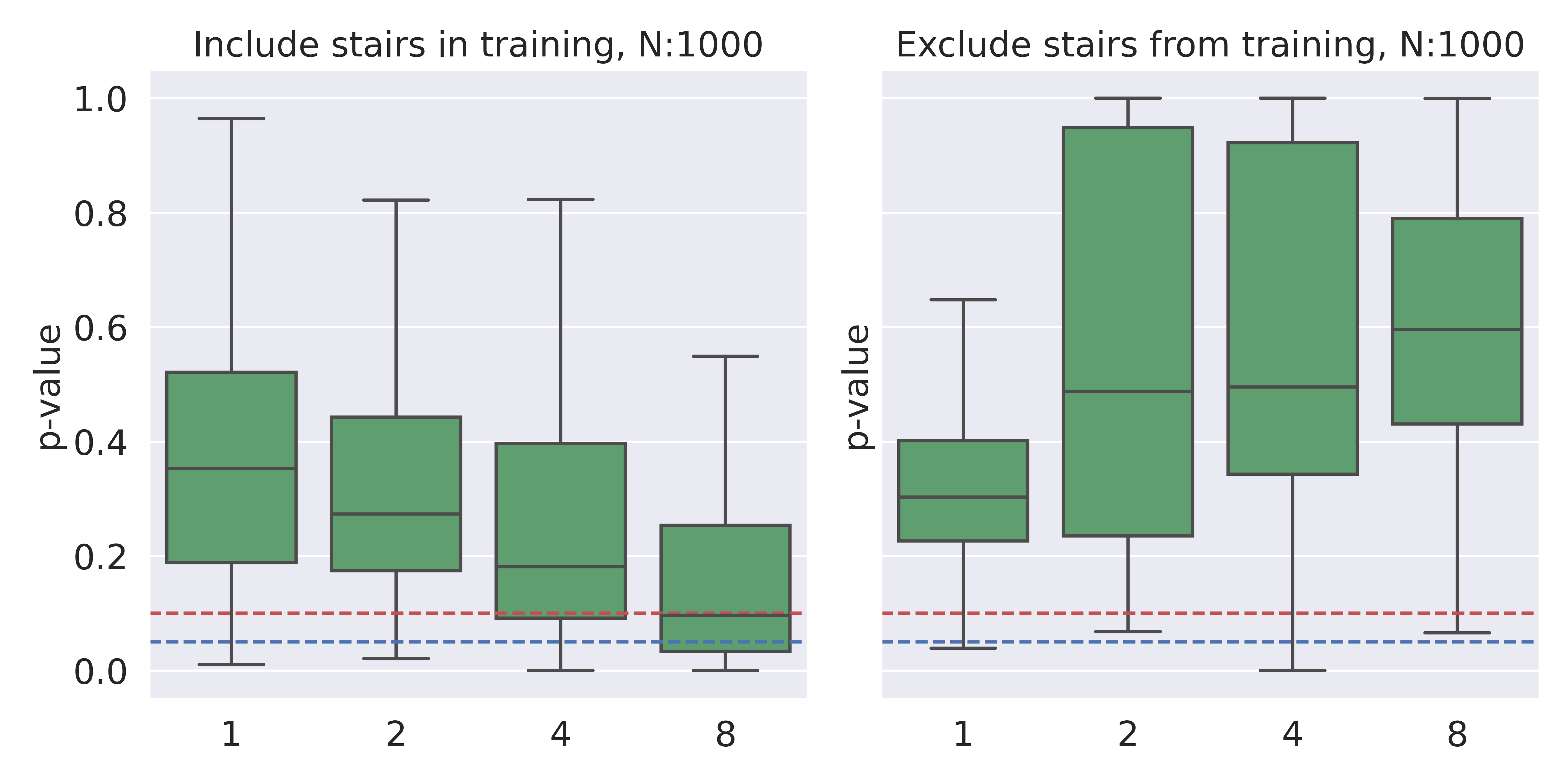}
    \caption{Nonparametric test on the smart home dataset.}
    \label{fig:nonparametric_hard_4975}
\end{figure}

In Figure \ref{fig:nonparametric_hard_4975}, the plot on the left corresponds to scenario (a), while scenario (b) is on the right. We do not expect there to be label noise in our walkaround dataset. Yet we observe that if we include RSSI readings received on stairs in training, p values will decrease as the number of anchor points increases, and finally go below the cut-off point. But if we exclude stairs from training, p values remain above the cut-off points, which says there is no class-conditional label noise in our walkaround dataset. This validates our assumption that grouping stairs into one of the floors might introduce noise in the labels. The stairs in our example house are wide and high enough, so that they will cause bias in training under setting (a), and finally lead us to wrong conclusions.


\section{Conclusions} 
In this work we showed how (nonparametric) hypothesis testing procedures on class-dependent label noise can be used when the underlying model is a product of local maximum likelihood estimation procedure. Going beyond the state-of-the-art, local Maximum Likelihood Estimation results in the creation of more adaptable nonparametric logistic regression models, thereby reducing vulnerability to model misspecification. This extends the potential applications to more occasions where the underlying data generation is unknown. We further compare empirically both parametric and nonparametric options, focusing on discussing
the advantages and limitations of the nonparametric approach. Finally, we presented a first real-world case study, demonstrating the steps of applying such tests in practice with a smart home dataset, which shows the effectiveness and practical considerations of the proposed method. 

In future work, we want to explore bootstrap bias correction as well as consider bootstrap on its own as a means of obtaining variance for predictions of anchors.

\bibliography{aaai24}

\clearpage
\onecolumn
\section{Appendix}

To further illustrate the behaviour of our proposed nonparametric tests, we present here additional experiments on synthetic data including several combinations of $(\alpha,~\beta)$, i.e. the difference in noise rates. In addition, we also run the baseline parametric tests proposed in \cite{poyiadzi2022statistical} under the same experiment settings for completeness.

\subsection{Symmetric XOR data: the role of $(\alpha,~\beta)$}
\begin{figure*}[h]
    \centering
    \includegraphics[width=0.5\linewidth]{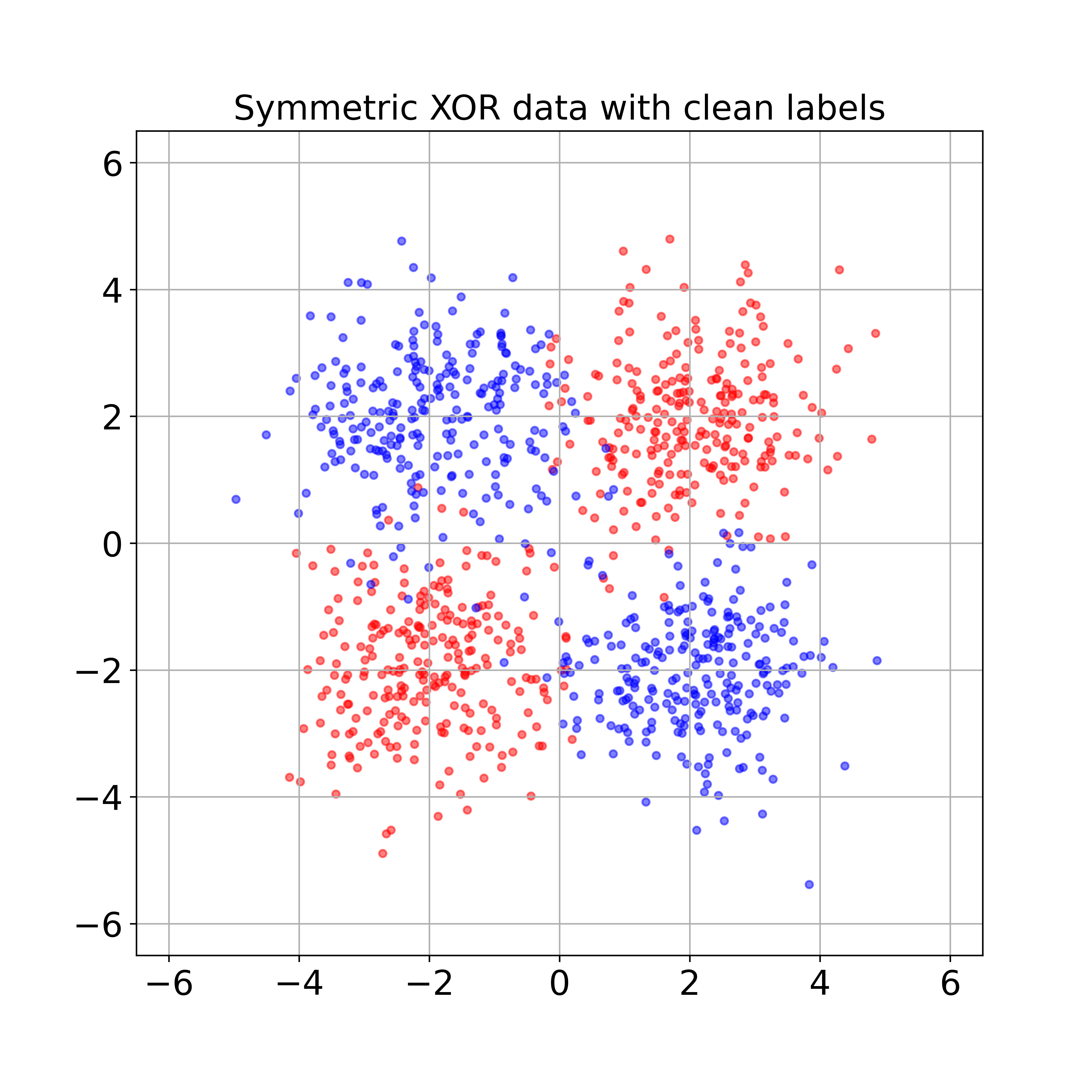}
    \caption{Symmetric XOR dataset}
    \label{fig:symmetricxor}
\end{figure*}

For the symmetric XOR data, the basic experiment settings are the same as in the experiments described in the main body of the paper. We consider four Gaussian distributions centred at: $[2, 2], [-2, -2], [-2, 2]$ and $[2, -2]$ with scale $1$ (first two correspond to one class, and the latter two to another), as shown in Figure \ref{fig:symmetricxor}. 

\vspace{+1cm}
Anchor points can be generated by first considering the optimal Bayes classifier (since we know the data generating distribution), and then optimising the input (the instance) such that the output is $0.50$. In the case of wanting relaxed anchor points, we do it with rejection sampling. In both cases, anchor points are restricted within $[-4, 4]$ for both features. We vary the following parameters:
\begin{enumerate}
    \item \label{enum:n} $N \in [200,~500,~1000]$: the training sample size.
    \item $k \in [1,~2,~4,~8,~16]$: the number of anchor points.\label{enum:k}
    \item $\delta \in [0,~0.05,~0.10,~0.20]$: how relaxed the anchor points are: $\eta(x) \in [0.50-\delta,~0.50+\delta]$.\label{enum:delta}
\end{enumerate}

\begin{figure*}[h]
    \centering
    \includegraphics[width=0.8\linewidth]{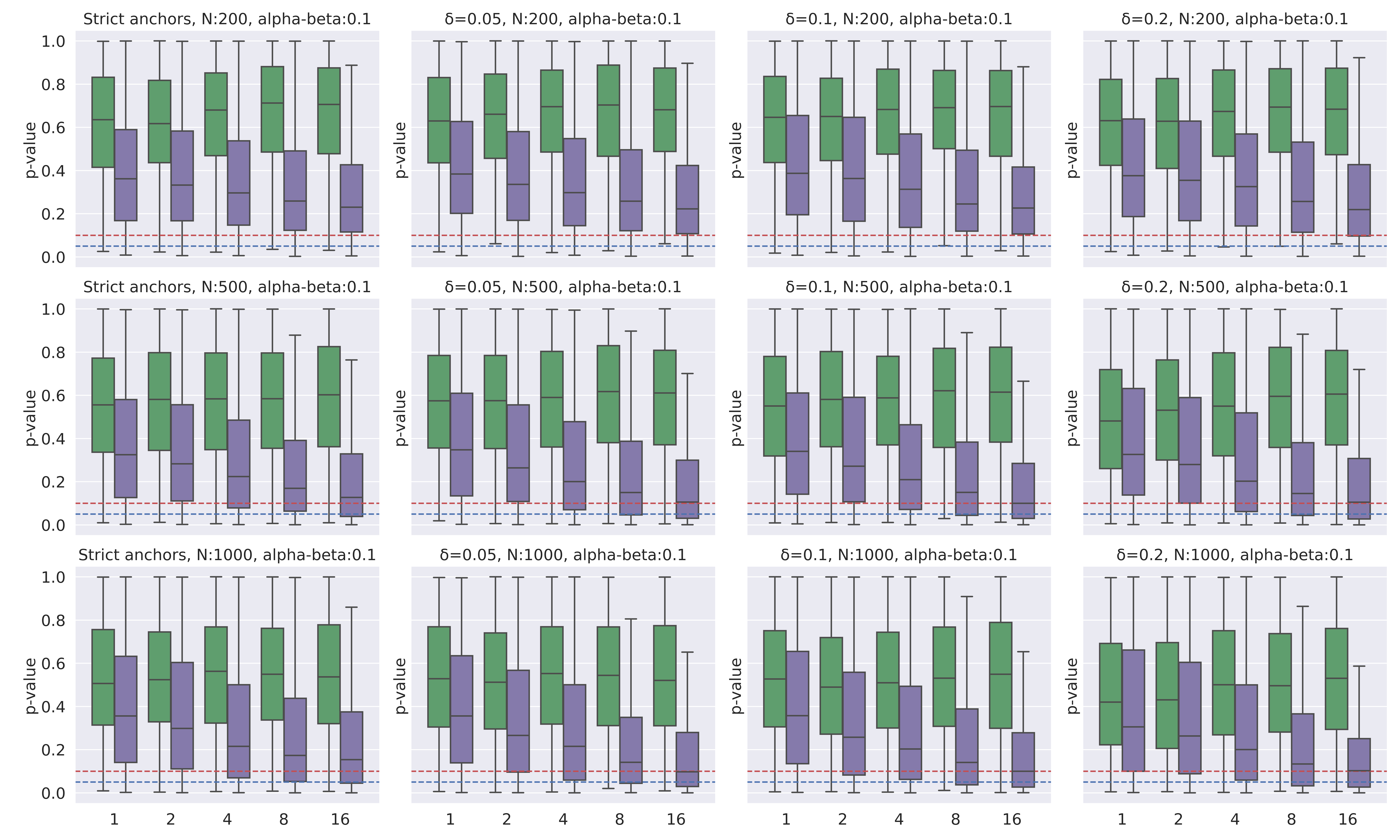}
    \caption{Nonparametric test on symmetric XOR data. Box-plots with $\alpha=0.2$ and $\beta=0.10$ (purple) against box-plots with $\alpha=0.0$ and $\beta=0.0$ (green).}
    \label{fig:nonparametric_alpha0.2_originalxor}
\end{figure*}

\begin{figure*}[!h]
    \centering
    \includegraphics[width=0.8\linewidth]{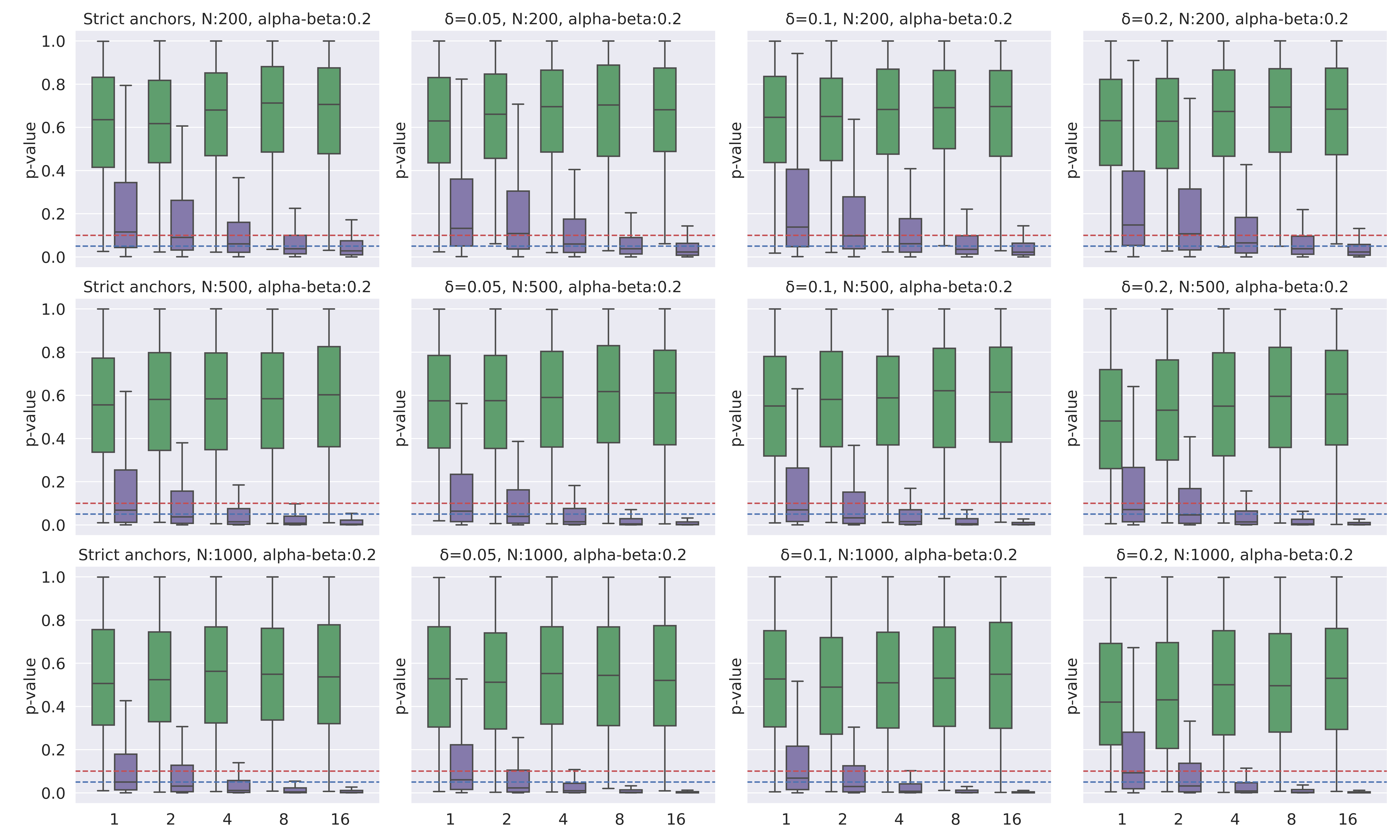}
    \caption{Nonparametric test on symmetric XOR data. Box-plots with $\alpha=0.3$ and $\beta=0.10$ (purple) against box-plots with $\alpha=0.0$ and $\beta=0.0$ (green).}
    \label{fig:nonparametric_alpha0.3_originalxor}
\end{figure*}

\begin{figure*}[h]
    \centering
    \includegraphics[width=0.8\linewidth]{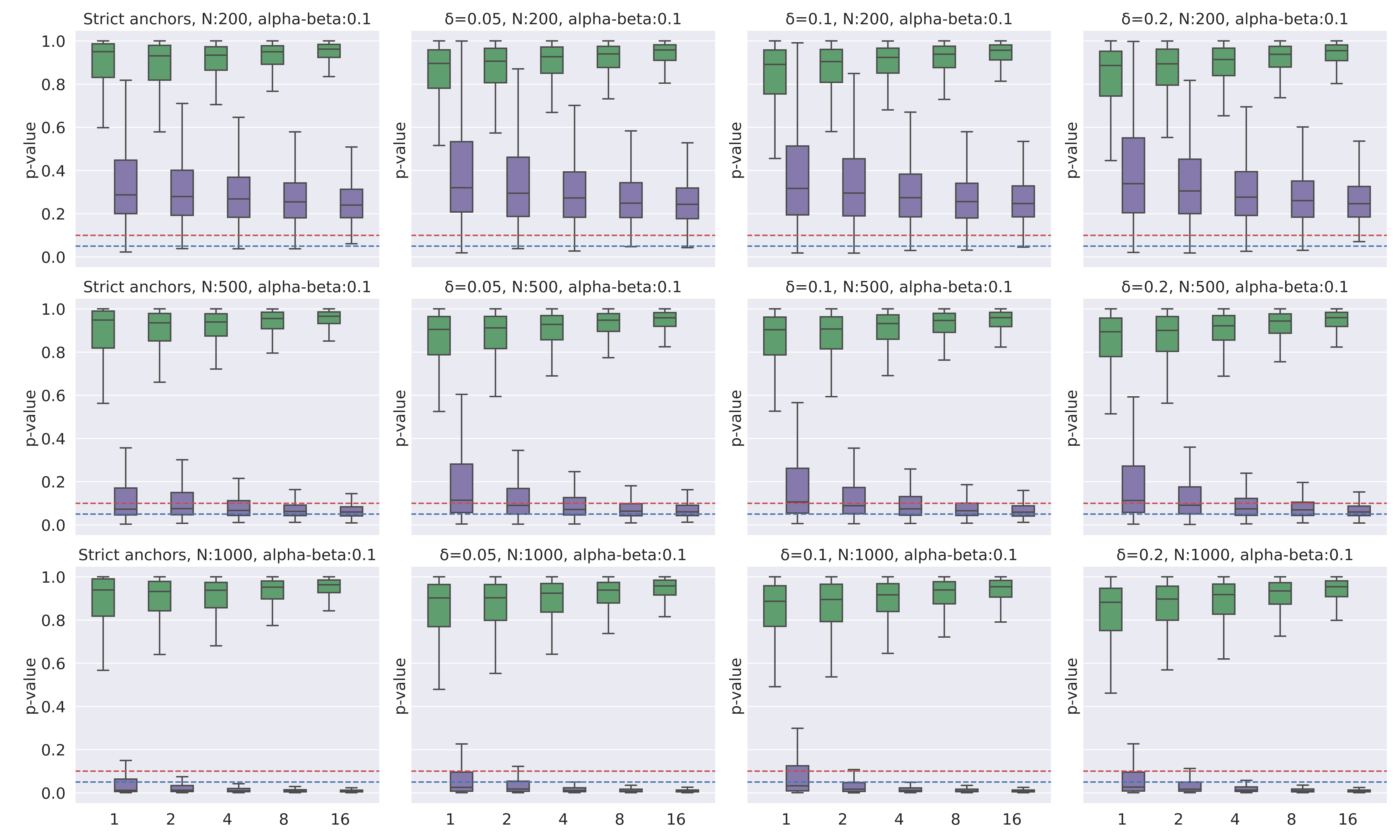}
    \caption{Parametric test on symmetric XOR data. Box-plots with $\alpha=0.2$ and $\beta=0.10$ (purple) against box-plots with $\alpha=0.0$ and $\beta=0.0$ (green).}
    \label{fig:parametric_alpha0.2_originalxor}
\end{figure*}

\begin{figure*}[h]
    \centering
    \includegraphics[width=0.8\linewidth]{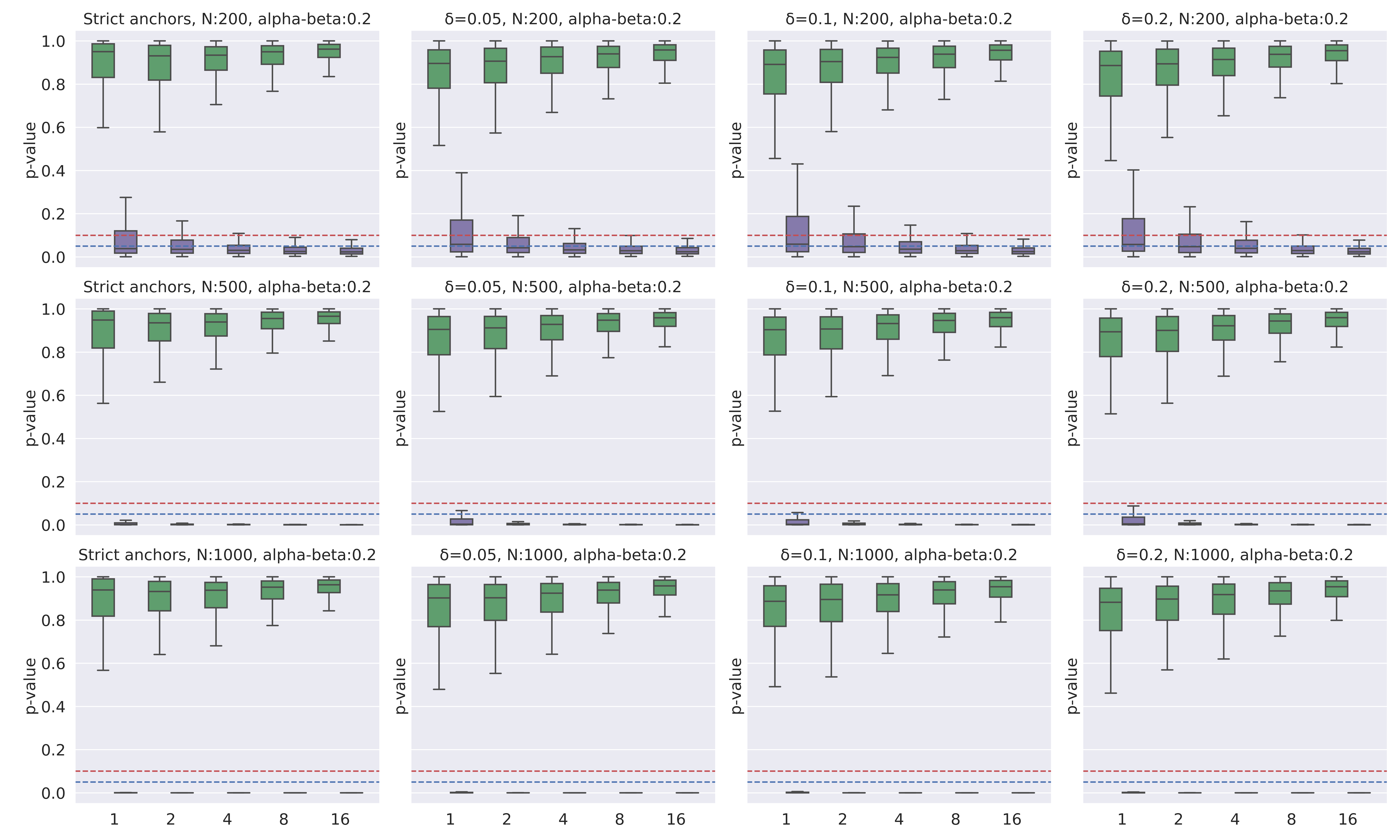}
    \caption{Parametric test on symmetric XOR data. Box-plots with $\alpha=0.3$ and $\beta=0.10$ (purple) against box-plots with $\alpha=0.0$ and $\beta=0.0$ (green).}
    \label{fig:parametric_alpha0.3_originalxor}
\end{figure*}

For each combination of $N$ and $(\alpha,~\beta)$ we perform $100$ runs from $100$ new dataset draws. Then for each run, and for each combination of $k$ and $\delta$ we do $10$ draws. We still fix $\beta=0.10$, but now we focus on varying $\alpha \in [0.2,~0.3]$. Figures
\ref{fig:nonparametric_alpha0.2_originalxor} and \ref{fig:nonparametric_alpha0.3_originalxor} are the results of our nonparametric test. Figures \ref{fig:parametric_alpha0.2_originalxor} and \ref{fig:parametric_alpha0.3_originalxor} are the results of the parametric test.

From Figures \ref{fig:nonparametric_alpha0.2_originalxor} and \ref{fig:nonparametric_alpha0.3_originalxor}, we can see that as the difference between noise rates $\alpha$ and $\beta$ increases, p-values correctly decrease, which is consistent with our previous observations. With respect to the results of the parametric test on this symmetric XOR data (Figures \ref{fig:parametric_alpha0.2_originalxor} and \ref{fig:parametric_alpha0.3_originalxor}), they can also be explained by the symmetry and overlap of the four Gaussians.

\clearpage

\subsection{Asymmetric XOR data: the role of $(\alpha,~\beta)$}
\begin{figure*}[h]
    \centering
    \includegraphics[width=0.5\linewidth]{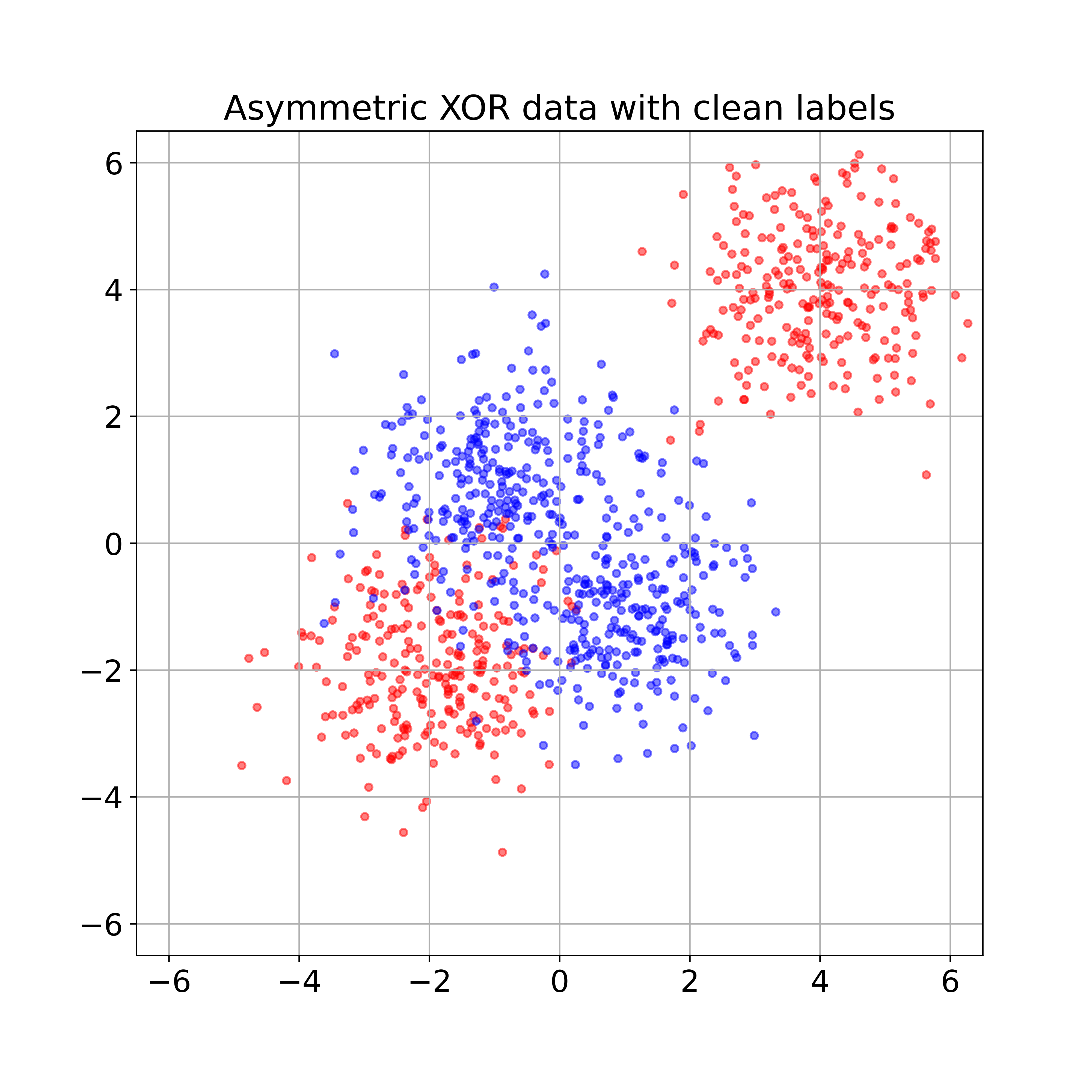}
    \caption{Asymmetric XOR dataset}
    \label{fig:asymmetricxor}
\end{figure*}

In the asymmetric XOR's case, the four Gaussians are centred at $[4, 4], [-2, -2], [-1, 1]$ and $[1, -1]$ with scale $1$ (first two correspond to one class, and the latter two to another), as Figure \ref{fig:asymmetricxor} shows.
\vspace{+1cm}

\begin{figure*}[h]
    \centering
    \includegraphics[width=0.8\linewidth]{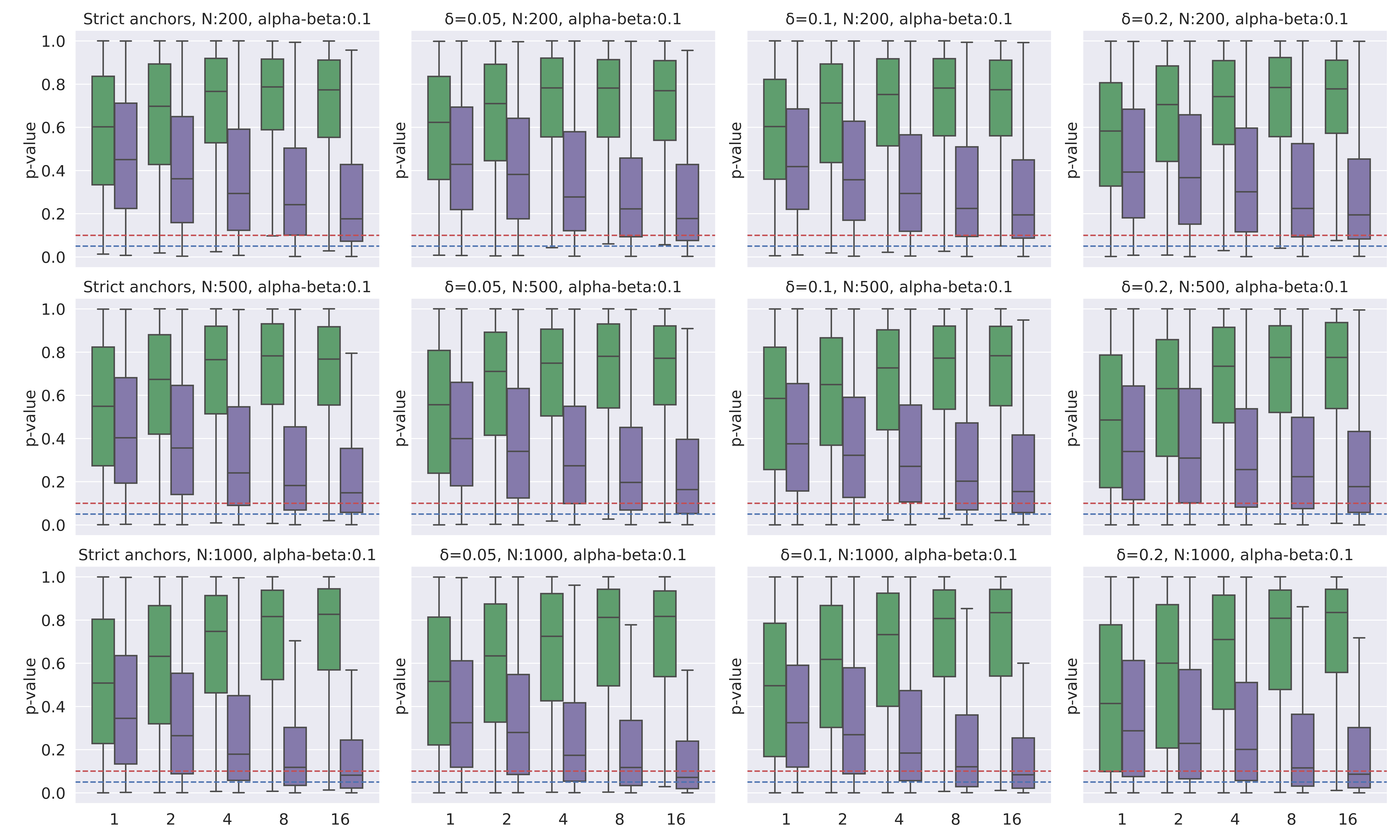}
    \caption{Nonparametric test on asymmetric XOR data. Box-plots with $\alpha=0.2$ and $\beta=0.10$ (purple) against box-plots with $\alpha=0.0$ and $\beta=0.0$ (green).}
    \label{fig:nonparametric_alpha0.2_d4xor2}
\end{figure*}

\begin{figure*}[h]
    \centering
    \includegraphics[width=0.8\linewidth]{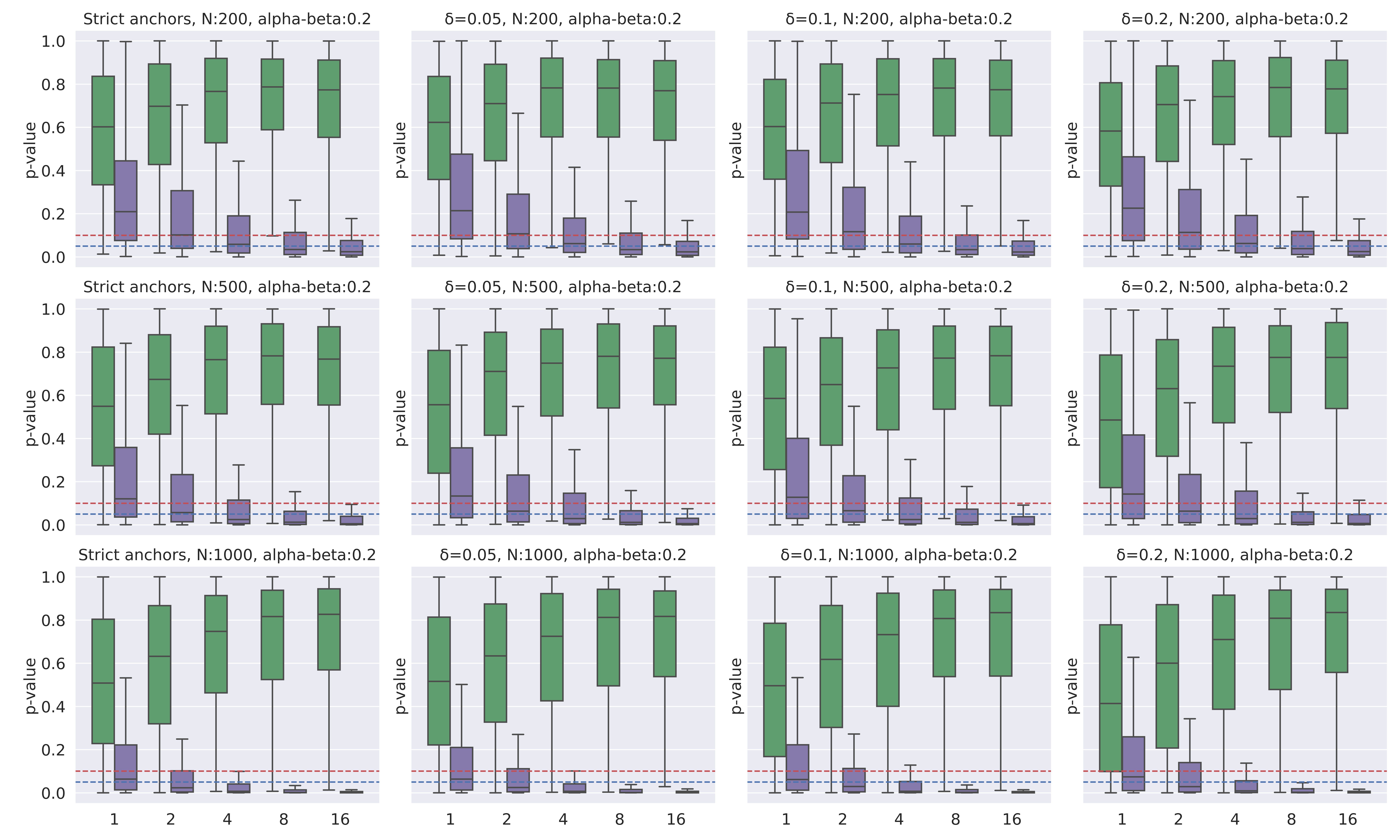}
    \caption{Nonparametric test on asymmetric XOR data. Box-plots with $\alpha=0.3$ and $\beta=0.10$ (purple) against box-plots with $\alpha=0.0$ and $\beta=0.0$ (green).}
    \label{fig:nonparametric_alpha0.3_d4xor2}
\end{figure*}

\begin{figure*}[h]
    \centering
    \includegraphics[width=0.8\linewidth]{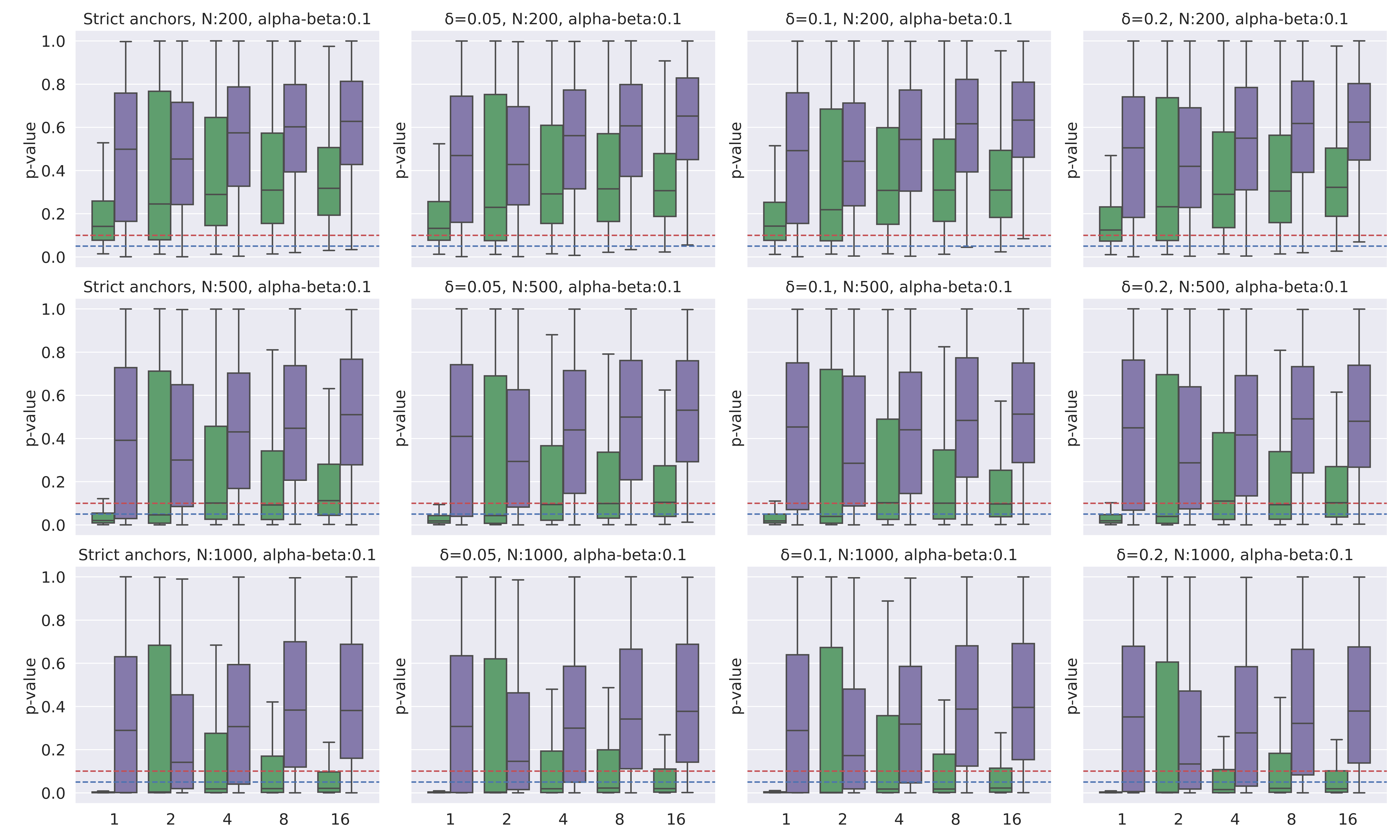}
    \caption{Parametric test on asymmetric XOR data. Box-plots with $\alpha=0.2$ and $\beta=0.10$ (purple) against box-plots with $\alpha=0.0$ and $\beta=0.0$ (green).}
    \label{fig:parametric_alpha0.2_d4xor2}
\end{figure*}

\begin{figure*}[h]
    \centering
    \includegraphics[width=0.8\linewidth]{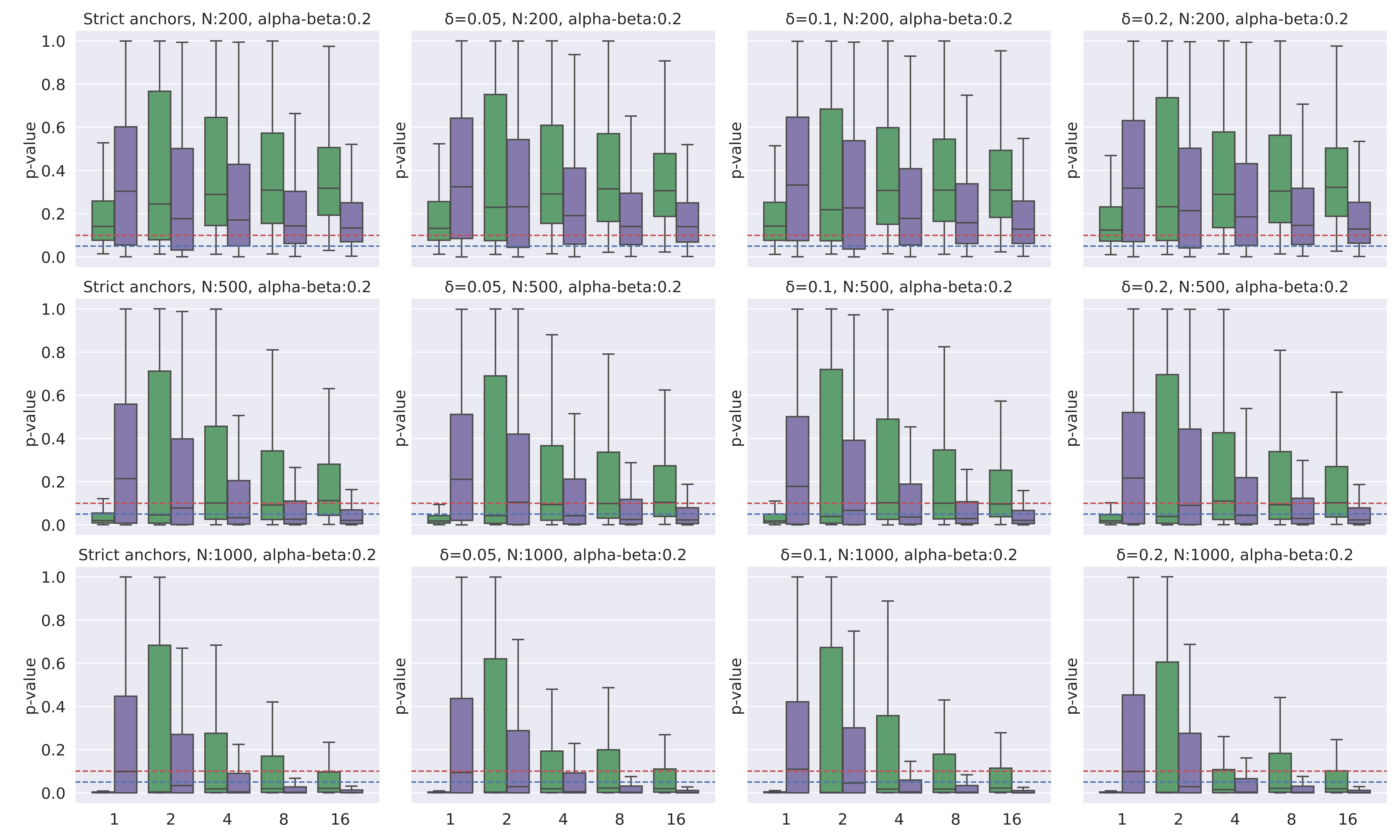}
    \caption{Parametric test on asymmetric XOR data. Box-plots with $\alpha=0.3$ and $\beta=0.10$ (purple) against box-plots with $\alpha=0.0$ and $\beta=0.0$ (green).}
    \label{fig:parametric_alpha0.3_d4xor2}
\end{figure*}

 We fix $\beta=0.10$, and we vary $\alpha \in [0.2,~0.3]$. The rest of the experiment settings are the same as those of the symmetric XOR dataset. Figures \ref{fig:nonparametric_alpha0.2_d4xor2} and \ref{fig:nonparametric_alpha0.3_d4xor2} correspond to the nonparametric test proposed in this paper, while Figures \ref{fig:parametric_alpha0.2_d4xor2} and \ref{fig:parametric_alpha0.3_d4xor2} are results of the parametric test.

 From Figure \ref{fig:nonparametric_alpha0.2_d4xor2} and \ref{fig:nonparametric_alpha0.3_d4xor2}, we can also observe that as the difference in noise rates increases, purple boxes, which correspond to the underlying model being trained on noisy data, will go down. At the same time, Figures \ref{fig:parametric_alpha0.2_d4xor2} and \ref{fig:parametric_alpha0.3_d4xor2} further confirm that the effectiveness of the parametric test depends on whether the assumptions of the model are met, which makes the parametric test less robust to model misspecification.


\end{document}